\documentclass[pdflatex,sn-mathphys-num]{sn-jnl}

\usepackage{graphicx}%
\usepackage{multirow}%
\usepackage{amsmath,amssymb,amsfonts}%
\usepackage{amsthm}%
\usepackage{mathrsfs}%
\usepackage[title]{appendix}%
\usepackage{xcolor}%
\usepackage{textcomp}%
\usepackage{manyfoot}%
\usepackage{booktabs}%
\usepackage{algorithm}%
\usepackage{algorithmicx}%
\usepackage{algpseudocode}%
\usepackage{listings}%
\usepackage{array}

\theoremstyle{thmstyleone}%
\theoremstyle{thmstyletwo}%

\theoremstyle{thmstylethree}%

\usepackage[english]{babel}
\usepackage[autostyle, english = american]{csquotes}
\MakeOuterQuote{"}

\begin{document}
	
	\title[Top 10 Open Challenges Steering the Future of Diffusion Language Model and Its Variants]{Top 10 Open Challenges Steering the Future of Diffusion Language Model and Its Variants}
	
	
	\author[1]{\fnm{Yunhe} \sur{Wang}}\email{yunhe.wang@huawei.com}	
	\author[1]{\fnm{Kai} \sur{Han}}
	\author[1]{\fnm{Huiling} \sur{Zhen}}
	\author[2]{\fnm{Yuchuan} \sur{Tian}}
	\author[1]{\fnm{Hanting} \sur{Chen}}
	\author[3]{\fnm{Yongbing} \sur{Huang}}
	\author[1]{\fnm{Yufei} \sur{Cui}}
	\author[2]{\fnm{Yingte} \sur{Shu}}
	\author[1]{\fnm{Shan} \sur{Gao}}
	\author[1]{\fnm{Ismail} \sur{Elezi}}
	\author[1]{\fnm{Roy Vaughan} \sur{Miles}}
	\author[1]{\fnm{Songcen} \sur{Xu}}
	\author[1]{\fnm{Feng} \sur{Wen}}
	\author[2]{\fnm{Chao} \sur{Xu}}
	\author[3]{\fnm{Sinan} \sur{Zeng}}
	\author[4]{\fnm{Dacheng} \sur{Tao}}
	
	\affil[1]{\orgname{Huawei Noah's Ark Lab}}
	\affil[2]{\orgname{Peking University}}
	\affil[3]{\orgname{Huawei Technologies}}
	\affil[4]{\orgname{Nanyang Technological University}}
	
	
	\abstract{The paradigm of Large Language Models (LLMs) is currently defined by auto-regressive (AR) architectures, which generate text through a sequential ``brick-by-brick'' process. Despite their success, AR models are inherently constrained by a causal bottleneck that limits global structural foresight and iterative refinement. Diffusion Language Models (DLMs) offer a transformative alternative, conceptualizing text generation as a holistic, bidirectional denoising process akin to a sculptor refining a masterpiece. However, the potential of DLMs remains largely untapped as they are frequently confined within AR-legacy infrastructures and optimization frameworks. In this Perspective, we identify ten fundamental challenges ranging from architectural inertia and gradient sparsity to the limitations of linear reasoning that prevent DLMs from reaching their ``GPT-4 moment''. We propose a strategic roadmap organized into four pillars: foundational infrastructure, algorithmic optimization, cognitive reasoning, and unified multimodal intelligence. By shifting toward a diffusion-native ecosystem characterized by multi-scale tokenization, active remasking, and latent thinking, we can move beyond the constraints of the causal horizon. We argue that this transition is essential for developing next-generation AI capable of complex structural reasoning, dynamic self-correction, and seamless multimodal integration.}

	
	
	
	\maketitle
	
	\section{Introduction to Diffusion Language Models}\label{sec1}
	The landscape of Natural Language Processing (NLP) has been fundamentally reshaped by the success of Large Language Models (LLMs)~\cite{brown2020language,achiam2023gpt,guo2025deepseek}, predominantly governed by the Auto-Regressive (AR) paradigm. Considering a sequence $\mathbf{x}=x^1,\cdots,x^N$, AR typically define the distribution as
    \begin{equation}
        \underbrace{p_\theta(x) = p_\theta(x^1) \prod_{n=2}^{N} p_\theta(x^n \mid x^1,\cdots,x^{n-1})}_{\text{Autoregressive LM formulation}}.
    \end{equation}    
    By decomposing the joint probability of a sequence into a product of conditional probabilities, AR models like the GPT series have demonstrated extraordinary capabilities in text generation and reasoning. However, this "left-to-right" sequential generation inherently suffers from several limitations: the accumulation of errors (exposure bias), a lack of global structural foresight, and the "causal bottleneck" that prevents the model from revising previous tokens based on future context.
	
	In parallel, Diffusion Models have emerged as the gold standard for continuous-domain generative tasks, such as high-fidelity image and video synthesis. Unlike AR models, diffusion models treat generation as a progressive denoising process—starting from a state of pure noise (or total masking) and iteratively refining the entire output toward a coherent structure. This is analogous to a sculptor, who begins with a rough block of marble and refines all parts of the statue simultaneously. When adapted to the discrete domain of language, Diffusion Language Models (DLMs)~\cite{nie2025large,ye2025dream,tian2025next} offer a compelling alternative: they enable non-sequential generation, support bidirectional context modeling, and allow for flexible "any-to-any" text editing and infilling. Let $\mathbf{x_t}$ deenote the noised sequence at the $t$-th timestep, the DLMs usually model the data distribution as
    \begin{equation}
        \underbrace{p_\theta(\mathbf{x}) = \sum_{\mathbf{x_{1:T}} \sim q} p(\mathbf{x_T}) \prod_{t=1}^{T} p_\theta(\mathbf{x_{t-1}} \mid \mathbf{x_t}),}_{\text{Diffusion LM formulation}}
    \end{equation}
    where $q$ is the forward process progressively corrupts the original data into noisy tokens.
	
	Despite their theoretical appeal, the transition of diffusion techniques from pixels to paragraphs has not been seamless. Text is inherently discrete, categorical, and highly structured, making the definition of "noise" and "denoising" far more complex than in continuous space. While early attempts at discrete diffusion and masked language modeling showed promise, they have yet to match the scaling efficiency and raw performance of their AR counterparts~\cite{ni2025diffusion,bie2025llada2}. Current DLMs often find themselves "trapped" in architectures and data pipelines originally optimized for AR tasks, leading to inefficiencies in inference, optimization, and structural reasoning. 

    While these limitations are often discussed in the context of text generation quality or efficiency, their impact becomes most evident in deep research and agentic scenarios, where models are required to maintain long-horizon goals, revise earlier hypotheses based on newly retrieved evidence, and iteratively refine structured outputs such as reports, plans, or theories. In such settings, the auto-regressive paradigm reveals fundamental cracks, whereas diffusion-based generation offers a more natural substrate for non-linear reasoning, global editing, and iterative belief revision.
	
    \begin{figure}[htp]
		\centering
		\includegraphics[width=1.0\textwidth]{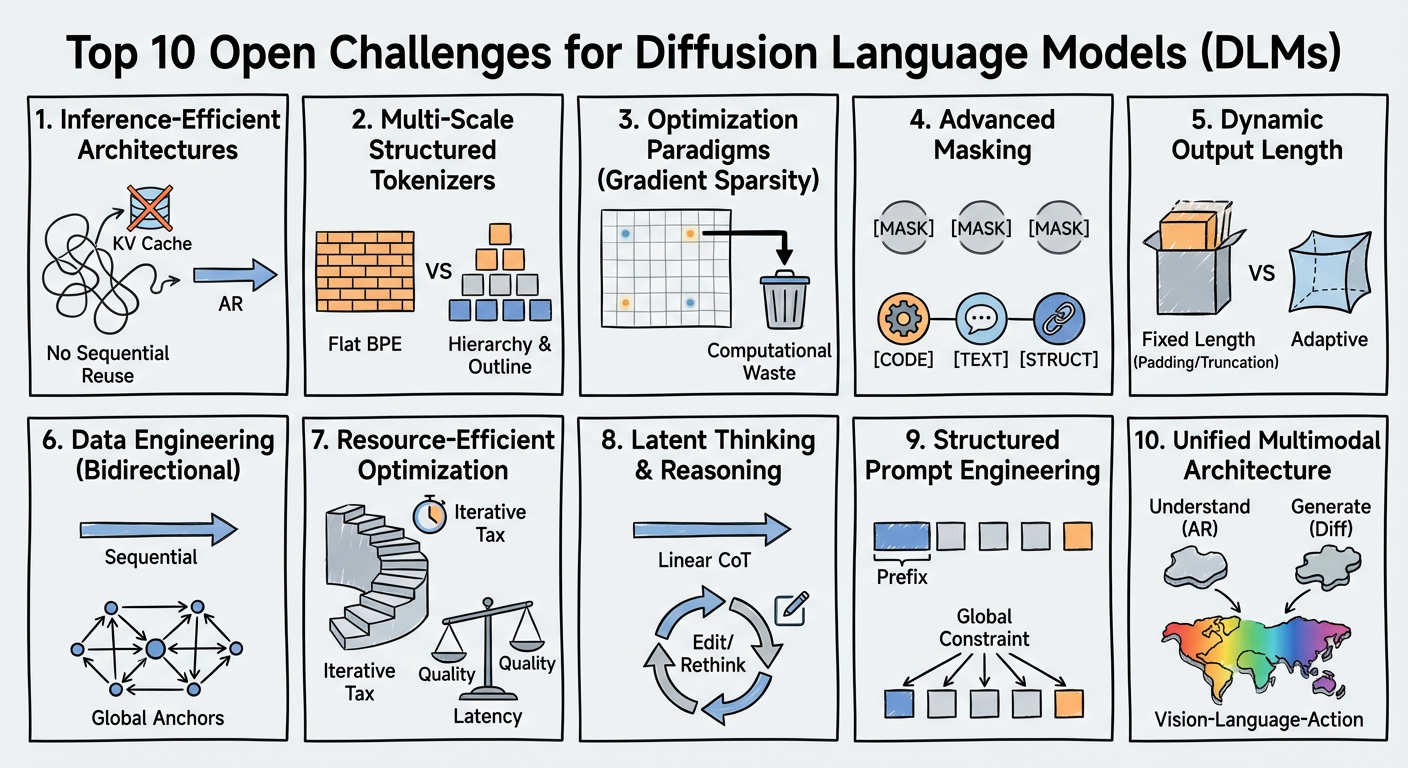}
		\caption{Illustration of Top 10 Open Challenges Steering the Future of Diffusion Language Model and Its Variants.}\label{fig:AI}
	\end{figure}
    
	This Perspective argues that for Diffusion Language Models to reach their full potential, we must move beyond merely adapting AR-centric frameworks and instead cultivate a native ecosystem designed for iterative, non-causal refinement. We identify ten critical dimensions that represent the frontier of this transition. We begin by rethinking the foundational infrastructure, advocating for inference-efficient architectures that transcend the limitations of traditional KV caching and multi-scale structured tokenizers that reflect the hierarchical nature of human thought. To address the inherent training inefficiencies of discrete diffusion, we propose novel optimization paradigms to mitigate gradient sparsity and advanced masking strategies that move beyond generic tokens toward functional, structured priors. Furthermore, we explore the necessity of dynamic output length mechanisms and diffusion-native data engineering to move away from the rigid constraints of fixed-length generation. To ensure practical viability, we discuss resource-efficient optimization techniques such as multi-step distillation and quantization tailored for denoising trajectories. Beyond mere generation, we envision a shift toward "slow thinking" through latent reasoning and structured prompt engineering, allowing models to utilize their bidirectional context for iterative editing and complex memory retrieval. Finally, we converge on the prospect of a unified architecture, \emph{i.e.}, a singular, end-to-end framework capable of harmonizing understanding, generation, and action across multiple modalities. By addressing these interconnected challenges, we can pave the way for a more robust, efficient, and structurally intelligent generation of AI that transcends the sequential constraints of the causal horizon.

\section{Top 10 Open Challenges}\label{sec2}

Despite the nascent potential of Diffusion Language Models, several fundamental bottlenecks remain. We identify ten core challenges that currently hinder their performance and scalability.

\paragraph{2.1) Inference-Efficient Architectures: Beyond AR Legacies}
The architectural backbone of most DLMs remains rooted in Transformer designs optimized for AR tasks. In AR models, the sequential nature of next-token prediction allows for efficient KV cache reuse~\cite{kwon2025vllm}. However, diffusion denoising is inherently non-sequential; the stochastic distribution of mask positions across iterations renders traditional KV caching mechanisms ineffective~\cite{wu2025fast,wu2025fast2}. Without a native architecture that supports bidirectional, iterative refinement without full-sequence re-computation, DLMs struggle to achieve the inference throughput necessary for large-scale deployment. This limitation becomes particularly acute for deep research agents, which require repeated global revisions of evolving artifacts (e.g., research drafts, hypotheses graphs). Without diffusion-native inference mechanisms, iterative agent loops become prohibitively expensive.

\paragraph{2.2) Structured Hierarchy in Tokenizers}
Current tokenization methods, such as Byte Pair Encoding (BPE)~\cite{sennrich2016neural}, are "flat" and statistically driven, lacking the structural hierarchy inherent in human cognition. While humans often conceptualize text through a multi-scale lens starting with global outlines before refining local details, existing DLMs are forced to operate on a uniform granularity. This mismatch prevents the model from efficiently allocating computational resources between high-level semantic structuring and fine-grained lexical polishing.

\paragraph{2.3) Optimization Paradigms: Addressing Gradient Sparsity}
A major source of inefficiency in DLM training is gradient sparsity. During long-sequence pre-training, the model is typically trained to denoise only a small, randomly masked subset of tokens within a long context (e.g., 32k tokens). Consequently, the vast majority of unmasked tokens in the forward pass do not contribute a loss, resulting in sparse and inefficient gradient feedback for the computational cost of a full forward and backward pass. This sparse supervision creates a second, downstream challenge: it introduces a distribution shift between pre-training (random masking) and downstream fine-tuning or reasoning, where the model must often generate or evaluate complete, coherent sequences (e.g., full answers in SFT or full reasoning traces in RL). This misalignment complicates the adaptation and alignment process.

\paragraph{2.4) Advanced Masking: From Generic to Structured Functionalism}
The prevailing "single [MASK] token" paradigm is elegant but functionally limited. By treating all masked positions as equivalent, it fails to account for the varying structural and logical importance of different tokens (e.g., a control flow operator in code vs. a filler word in prose). The current masking approach lacks a structured mechanism that considers the interdependencies between masked positions, leading to a lack of functional diversity in the model's restoration capabilities. Furthermore, considering a deep research agent, masking a factual citation and masking a core logical claim should trigger fundamentally different restoration behaviors—something the current generic [MASK] paradigm cannot express.

\paragraph{2.5) Dynamic Output Length and Adaptive Termination}
Unlike AR models that naturally terminate via an End-of-Sequence (EOS) token, DLMs typically require a pre-defined output length. This rigidity is computationally inefficient: tasks requiring minimal reasoning may be forced into long-sequence windows, while complex tasks may suffer from premature truncation. Current methods struggle to adaptively infer the optimal length for a given query, leading to either "hallucinatory padding" or information loss.

\paragraph{2.6) Data Engineering: Curating for Bidirectional Learning}
Most DLMs are trained on data curated for AR models, which emphasizes sequential continuity. However, to unlock the full potential of bidirectional denoising, models require data that highlights structural relationships and multi-point dependencies. Current datasets do not explicitly support the learning of global semantic "anchors," making it difficult for DLMs to develop the same level of structural intelligence as they do in the continuous image domain.

\paragraph{2.7) Resource-Efficient Model Optimization}
While DLMs offer theoretical parallel generation, the "iterative tax" of multiple denoising steps often results in higher latency than AR models at parity. When batch sizes increase, the global attention overhead of diffusion can negate its speed advantages~\cite{fu2025bits}. Finding the right balance between denoising quality and computational cost remains an open challenge, particularly when basic structural components of DLMs have not yet converged.

\paragraph{2.8) Latent Thinking and Iterative Reasoning}
Reasoning in LLMs is often equated with sequential Chain-of-Thought (CoT)~\cite{wei2022chain,bi2025forest}. For DLMs, simply mimicking this linear path is suboptimal. Current SFT paradigms fail to leverage the model's ability to "re-think" or "edit" its output during the denoising process. If the model is forced into a predetermined length space without a mechanism for deep, latent refinement, it cannot effectively perform the iterative self-correction that characterizes complex human reasoning. Moreover, deep research is inherently non-linear: hypotheses are proposed, invalidated, and reformulated. Diffusion-native latent thinking provides a natural mechanism for such iterative belief revision, whereas linear Chain-of-Thought enforces an unnatural reasoning trajectory.

\paragraph{2.9) Structured Prompt Engineering and Contextual Memory}
Traditional prefix-based prompting is a byproduct of causal modeling. For a bidirectional DLM, the prompt could theoretically be interleaved with the generation or serve as a global constraint. However, we currently lack a standardized framework for "Diffusion-Native Prompting". This limits the model's effectiveness in scenarios like Deep Research or Agentic tasks, where a few global key tokens should ideally trigger a full-sequence logical reconstruction.

\paragraph{2.10) Toward a Unified Multimodal Architecture}
The field is currently fragmented: "understanding" tasks typically rely on AR architectures, while "generation" favors Diffusion. In complex domains like Vision-Language-Action (VLA)~\cite{black2024pi_0}, this leads to hybrid models where different modalities are governed by different optimization goals. Achieving a truly unified architecture where understanding, generation, and action are treated as points on a continuous diffusion spectrum remains the ultimate frontier.

\section{Strategic Insights for the Challenges}\label{sec3}

To overcome the aforementioned bottlenecks, we propose a strategic roadmap focused on shifting from "AR-adaptation" to a "Diffusion-native" ecosystem. These insights are categorized into four pillars: architectural foundations, optimization mechanics, cognitive reasoning, and the path to a unified intelligence.

\subsection{Pillar I: Infrastructure and Structural Foundations}

\textbf{Redesigning for Non-Causal Efficiency.} We advocate for attention structures natively designed for diffusion, such as stochastic-aware attention or partial KV caching tied to specific masking patterns. While enforcing a left-to-right denoising order offers a temporary efficiency gain, the long-term solution lies in a fundamental redesign of the KV structure that can handle non-causal, iterative updates without redundant global re-computation.

\noindent\textbf{The Multi-scale Tokenizer Framework.} To reflect the hierarchical nature of human language, we propose a multi-scale tokenization approach. By employing multi-layered vocabularies where high-level tokens represent paragraph-level semantic bridges (outlines) and low-level tokens handle fine-grained lexical details, the model can simulate hierarchical thought. This requires a shift in the training-inference paradigm to accommodate multi-resolution data streams, allowing the model to "sculpt" global structure before "filling" local content.

\subsection{Pillar II: Algorithmic Mechanics and Optimization}

\textbf{Dynamic Optimization and Efficiency.} To mitigate gradient sparsity, we suggest dynamic masking ratios that evolve across training stages (e.g., starting with high-ratio global masking and moving to low-ratio local refinement). High-impact directions for inference efficiency include multi-step trajectory distillation (reducing denoising steps to $N < 5$) and speculative decoding~\cite{gao2025self}, where a smaller model proposes a draft for a larger model to refine. Additionally, hybrid regimes that use DLMs for long-sequence planning and AR for high-throughput execution could offer a synergistic computational balance.

\noindent\textbf{Functional Masking and Elastic Generation.} We advocate for "Structured Masking" using multiple specialized mask tokens (e.g., [LOGIC-MASK], [ENTITY-MASK]) with prior-informed dependencies. By introducing a functional hierarchy to masking, DLMs can better capture complex structural constraints. Furthermore, incorporating EOS-position prediction directly into the denoising steps allows the model to perceive the optimal output length dynamically, enabling an "elastic" generation window that adaptively terminates based on the query's complexity~\cite{li2025fixedtrainingfreevariablelengthdenoising,shu2026deferredcommitmentdecodingdiffusion}.

\subsection{Pillar III: Cognitive Reasoning and Interaction}

\textbf{Diffusion-Native CoT.} We propose shifting from linear, sequential reasoning to an iterative "outline-then-detail" process. During denoising, the model should engage in Active Remasking: identifying low-confidence tokens or logical inconsistencies and "re-masking" them for immediate re-generation. This internal feedback loop enables a form of "latent thinking" and self-correction that surpasses the rigid, forward-only limits of traditional sequential logic.

\noindent\textbf{Scaffolding Prompts and Contextual Memory.} We suggest transitioning from "Prefix-Prompts" to "Cloze-Scaffolding". By providing global anchor tokens as a skeletal prompt interleaved throughout the sequence, DLMs can perform rapid, high-fidelity infilling. This paradigm is particularly suited for RAG and long-term memory management, where the prompt acts as a non-sequential retrieval cue that guides the global denoising trajectory.

\subsection{Pillar IV: Data Engineering and Unified Intelligence}

\textbf{Diffusion-Native Data Ecosystems.} Data engineering must move toward curating "structural dependencies" rather than just "sequences." This involves annotating "anchor tokens" and structural landmarks within pre-training corpora to guide the model's focus. Structured SFT and RL datasets should be redesigned to emphasize multi-point editing and bidirectional restoration, rewarding the model for global coherence and logical stability.

\noindent\textbf{The Path to a Unified Diffusion Backbone.} The ultimate frontier is a Unified Diffusion Objective that treats understanding (high-noise denoising) and generation (low-noise denoising) as part of a single continuum. In Vision-Language-Action (VLA) models, this would allow for a seamless transition between perceiving the environment and executing actions, using a singular, natively multimodal backbone that collapses the modality gap into a unified denoising manifold.

\subsection{Pillar V: DLMs as the Cognitive Core of AI Agents}

Rather than introducing an additional pillar, we view deep research agents as a systems-level instantiation of the four pillars discussed above. Deep research is not merely an extended form of text generation, but a cognitive process characterized by hypothesis formation, evidence aggregation, contradiction resolution, and the iterative refinement of structured artifacts. Unlike single-pass generation, research inherently requires the ability to revisit, revise, and globally reorganize earlier conclusions in light of newly acquired information. From this perspective, diffusion language models provide a more compatible generative substrate for deep research agents, as their denoising-based formulation naturally supports non-linear revision and global structural editing.

In practice, a deep research agent may repeatedly re-evaluate and modify an evolving research draft, hypothesis graph, or literature synthesis as new evidence is retrieved or prior assumptions are invalidated. Diffusion-native architectures that support partial updates, structured masking, and inference-efficient global refinement render such long-horizon research loops computationally viable~\cite{anonymous2025beyond}. In contrast, auto-regressive generation typically enforces full sequential regeneration, leading to unnecessary recomputation and limiting the feasibility of iterative, research-style workflows. More importantly, diffusion-based latent thinking enables a form of reasoning that aligns closely with the epistemic dynamics of research. Rather than committing to a fixed linear chain of thought, the model can identify low-confidence, contradictory, or logically fragile regions within its own output and actively re-mask them for targeted denoising. This mechanism enables iterative self-correction and internal consistency checking, resembling a form of implicit peer review that is difficult to realize under strictly causal generation paradigms.

Viewed through this lens, deep research agents are not a downstream application layered on top of diffusion language models, but a natural expression of their core inductive bias. By unifying global planning, local refinement, and iterative belief revision within a single denoising framework, diffusion language models offer a principled foundation for research-oriented, agentic intelligence.

\section{Conclusion}\label{sec4}

The transition from the current auto-regressive dominance to a more balanced or unified diffusion-based landscape represents more than a mere shift in generative modeling; it is a fundamental move toward a more robust and human-like intelligence. As we have argued, the "causal horizon" of next-token prediction, while powerful, imposes intrinsic limits on structural foresight, error correction, and multi-scale reasoning. To transcend these boundaries, the research community must move beyond the "AR-adaptation" phase and embrace a native ecosystem for text diffusion.

Addressing the ten open challenges identified in this Perspective—ranging from hardware-aware non-causal architectures to the realization of latent, non-linear thinking—requires a concerted effort across the stack of model design. It demands a fundamental rethink of how we discretize language via tokenization, how we optimize for gradient efficiency, and how we interact with models that can "re-think" and "edit" their own internal trajectories. By cultivating these "diffusion-native" principles, we can develop architectures that are not only more computationally flexible and efficient but also inherently capable of the global structural reasoning that has remained elusive for purely sequential models.

Ultimately, the path to a unified intelligence lies in collapsing the divide between understanding, generation, and action. Diffusion models, with their unique ability to treat these tasks as different facets of a singular denoising manifold, offer a promising blueprint for this unification. As we solve the bottlenecks of inference latency and optimization stability, Diffusion Language Models may well become the cornerstone of a new era of AI that does not just predict the next word, but scrupulously scuplts the entire structure of thought.

%
%
	
	
	\bibliography{sn-bibliography}

\end{document}